\documentclass[letterpaper]{article} 
\usepackage{aaai2026}  
\usepackage{times}  
\usepackage{helvet}  
\usepackage{courier}  
\usepackage[hyphens]{url}  
\usepackage{graphicx} 
\usepackage{natbib}  
\usepackage{caption} 
\usepackage{amsmath,amssymb,amsfonts}
\usepackage{algorithm}
\usepackage{algorithmic}
\newcommand{\scheme}{Protego}
\usepackage{multirow}
\usepackage{hhline}
\usepackage{subcaption}
\usepackage{xcolor}
\usepackage{bbding}

\usepackage{newfloat}
\usepackage{listings}
\DeclareCaptionStyle{ruled}{labelfont=normalfont,labelsep=colon,strut=off} 
\floatstyle{ruled}
\newfloat{listing}{tb}{lst}{}
\floatname{listing}{Listing}
\nocopyright

\title{Protego: User-Centric Pose-Invariant Privacy Protection Against\\Face Recognition-Induced Digital Footprint Exposure}
\author{
    Ziling Wang, Shuya Yang, Jialin Lu, Ka-Ho Chow\textsuperscript{\Envelope}}
\affiliations{
    School of Computing and Data Science\\
    The University of Hong Kong
}

\begin{document}

\maketitle

\begin{abstract}
Face recognition (FR) technologies are increasingly used to power large-scale image retrieval systems, raising serious privacy concerns. Services like Clearview AI and PimEyes allow anyone to upload a facial photo and retrieve a large amount of online content associated with that person. This not only enables identity inference but also exposes their digital footprint, such as social media activity, private photos, and news reports, often without their consent. In response to this emerging threat, we propose \scheme{}, a user-centric privacy protection method that safeguards facial images from such retrieval-based privacy intrusions. \scheme{} encapsulates a user’s 3D facial signatures into a pose-invariant 2D representation, which is dynamically deformed into a natural-looking 3D mask tailored to the pose and expression of any facial image of the user, and applied prior to online sharing. Motivated by a critical limitation of existing methods, \scheme{} amplifies the sensitivity of FR models so that protected images cannot be matched even among themselves. Experiments show that \scheme{} significantly reduces retrieval accuracy across a wide range of black-box FR models and performs at least $2\times$ better than existing methods. It also offers unprecedented visual coherence, particularly in video settings where consistency and natural appearance are essential. Overall, \scheme{} contributes to the fight against the misuse of FR for mass surveillance and unsolicited identity tracing.
\end{abstract}

\section{Introduction}
\begin{figure}\centering
	\includegraphics[width=0.925\linewidth]{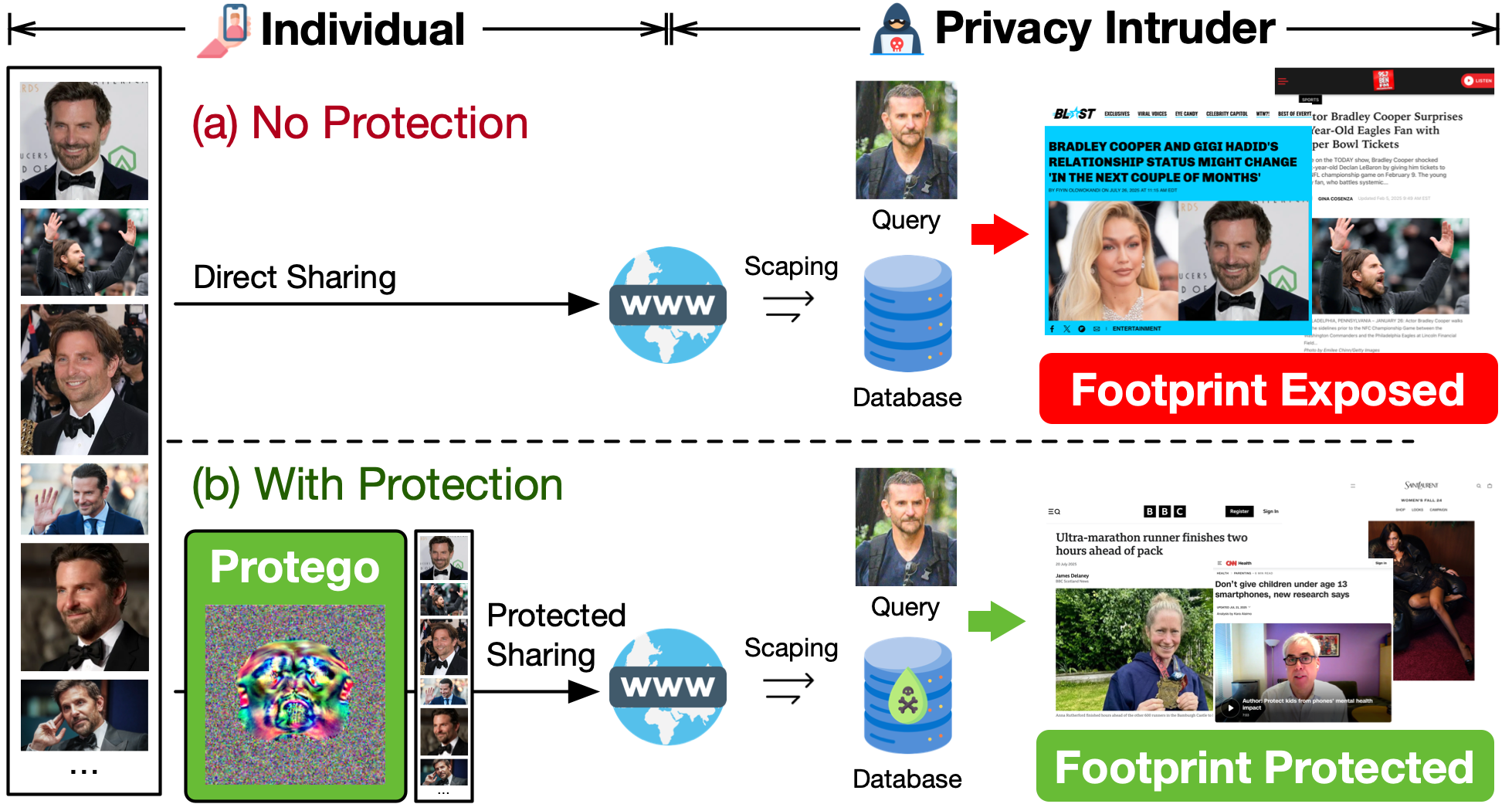}
	\caption{(a) A privacy intruder can scrape online platforms to build a database and retrieve an individual’s digital footprint using a facial image as a query. (b) \scheme{} protects users by enabling effective obfuscation of both images and videos before sharing. Query images, whether protected or not, fail to yield correct matches in the intruder’s database.}\label{fig:showcase}
\end{figure}

While offering many life-enriching applications, face recognition (FR) can be misused to invade personal privacy~\cite{wang2024beyond}. A growing concern is the use of FR to power large-scale search engines that retrieve online content based on facial images. Companies like \citet{clearview} and \citet{pimeyes}  have scraped billions of photos from the Internet to construct massive databases (DBs), spanning millions of individuals worldwide~\cite{Daniel}. By uploading a photo, anyone can uncover a vast collection of web content that contains the same person. As shown in Figure~\ref{fig:showcase}a, this retrieval not only enables identity inference but also exposes a person's digital footprint, such as social media activity, personal photos, and news reports. While some jurisdictions have demanded the removal of their citizens’ data~\cite{Laurie}, the lack of global regulatory consensus continues to hinder enforcement~\cite{Lydia}. \emph{Can individuals proactively safeguard their future digital footprint?}
\begin{figure*}\centering
\includegraphics[width=0.93\linewidth]{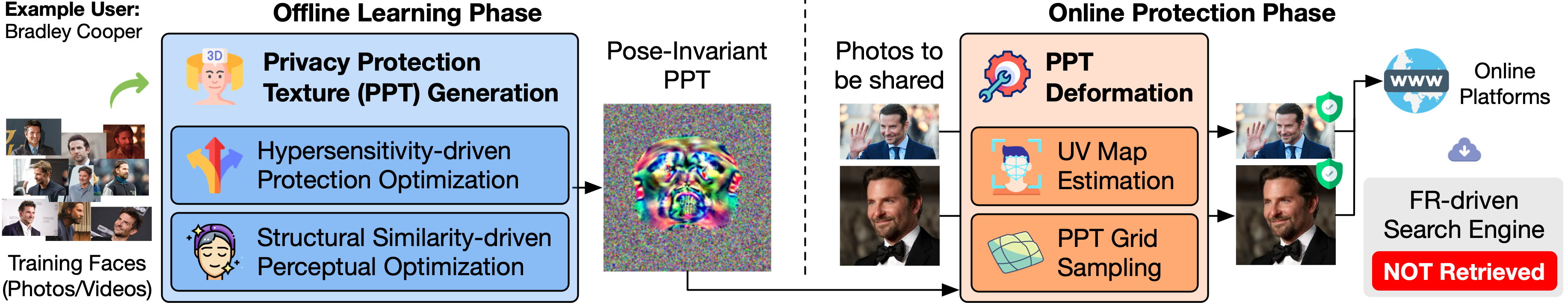} 
\caption{\scheme{} generates a user-specific, pose-invariant PPT during an offline learning phase. This PPT can be applied to any facial image of the user by deforming it to match the input’s pose and expression.}\label{fig:overview}
\end{figure*}

The need for such protective measures has led to the development of anti-FR technologies~\cite{wenger2023sok}. They work by subtly altering facial images before they are shared online. The goal is to prevent these protected images from being retrieved if they are later scraped and included in the privacy intruder's DB. These techniques are designed with two key objectives. First, they aim to significantly change the facial features extracted by FR models so that protected images cannot be matched to their original, unprotected versions. Second, the visual changes must remain minimal to preserve the appearance of the image. 

While many attempts have been made~\cite{yang2021towards,zhong2022opom,chow2024diversity,chow2024personalized}, existing methods are still fragile on two fronts. From an effectiveness perspective, they assume that privacy intruders will only query the system using unprotected facial images. However, if a protected image is used as a query, the system can retrieve the corresponding protected entries from the DB with high accuracy. This scenario is highly plausible, as query images can come from various sources. For example, an intruder might download a knowingly or unknowingly protected facial image from the Internet, use it as a query, and uncover the victim’s digital footprint to gather more information about them. A second common limitation lies in the visual assumptions: these methods expect faces to be front-facing, which is an unrealistic constraint, especially in video settings where head pose naturally varies over time.

To address the above gaps, we propose \scheme{} for better privacy protection and visual naturalness  (Figure~\ref{fig:showcase}b). \scheme{} learns from a small number of facial images of the user, encoding their 3D facial signatures into a pose-invariant 2D representation called the privacy protection texture (PPT). The PPT is trained using a novel loss that makes FR models hypersensitive to protected content, ensuring that protected images yield significantly different features and cannot be matched, even among themselves. At deployment, the PPT is dynamically deformed into a natural-looking 3D mask that aligns with the pose and expression of any facial image of the user, and is applied prior to sharing the image online.

In summary, we make the following contributions. First, we reveal two fundamental limitations shared by all existing methods that compromise both protection and user experience. Second, we introduce a novel hypersensitivity loss that ensures robust protection even when the query image itself has been protected. Third, we propose a new 3D protection paradigm that delivers natural-looking, pose-aware protection while remaining computationally efficient. Extensive experiments demonstrate that \scheme{} achieves at least twice the protection performance of state-of-the-art methods, with high-quality results in videos and strong generalization across unseen FR models.

\section{Background}
\subsection{FR-based Search Engine}
To build a search engine that retrieves content based on facial images, the owner first constructs a database $\boldsymbol{\mathcal{D}}$, where each entry contains at least a facial image that serves as a search index. A pretrained FR model $F$ is then used to extract features $F({\boldsymbol{x}})$ from each image ${\boldsymbol{x}}\in\boldsymbol{\mathcal{D}}$. When a facial image $\boldsymbol{x}^q$ is provided as a query, the same FR model is applied to obtain its features $F(\boldsymbol{x}^q)$, which are then compared to those of the DB entries. The search results are defined as:
\begin{equation}
	\textsc{Search}(\boldsymbol{x}^q;F,\boldsymbol{\mathcal{D}})=\underset{\tilde{\boldsymbol{x}}\in\boldsymbol{\mathcal{D}}}{\textsc{Top}_K}\big[\textsc{Sim}(F(\boldsymbol{x}^q), F(\tilde{\boldsymbol{x}}))\big],
\end{equation}
where $\textsc{Top}_K$ is an operator that returns the top-$K$ most similar entries and $\textsc{Sim}$ measures the cosine similarity.

\subsection{Related Work}
Existing anti-FR approaches can be broadly categorized into synthesis-based~\cite{zhu2019generating,deb2020advfaces,hu2022protecting} and perturbation-based methods~\cite{shan2020fawkes,cherepanova2021lowkey,yang2021towards,zhong2022opom,chow2024diversity,chow2024personalized}. This work focuses on the latter, which applies subtle changes to facial images to better preserve the user’s visual identity~\cite{wenger2023sok}.
Fawkes~\cite{shan2020fawkes} pioneered an untargeted attack that pushes facial features away in the embedding space. LowKey~\cite{cherepanova2021lowkey} improved robustness by modeling the entire image processing pipeline. PMask~\cite{chow2024diversity} enhances protection against unknown FR models via ensemble diversity. However, these methods rely on per-image iterative optimization, which can take minutes even on a GPU.
To improve efficiency, OPOM~\cite{zhong2022opom} and Chameleon~\cite{chow2024personalized} propose learning a single perturbation mask per user, enabling rapid protection of any image from the same individual. However, these masks expect faces to be front-facing, which limits their effectiveness in real-world scenarios such as videos. A more critical limitation shared across all existing methods is that their protected images have highly similar features, which undermines their effectiveness when both the query and the DB entries are protected. In contrast, \scheme{} preserves the strengths of prior methods while addressing these challenges by offering efficient, pose-robust protection that remains effective even when the query image is protected. 

\subsection{Threat Model}
We consider a threat model in which a privacy intruder performs large-scale web scraping to construct a DB. Each entry in the DB consists of a person's facial image and a snapshot of the web content from which it was obtained. When given a query image, the intruder uses an FR model to search for matching entries and returns the associated web content, allowing sensitive information about the person to be revealed. The goal of \scheme{} is to enable users to preprocess their photos before sharing them online, so that even if the images are scraped and stored in the DB, they will not be retrieved given a query of the same individual. Unlike existing methods, we consider a more realistic and challenging setting in which the query image may also be protected.

\section{Methodology}
Figure~\ref{fig:overview} illustrates the overall workflow of \scheme{}, which consists of two phases: (i) an offline learning phase that encapsulates a user's facial signatures into a PPT, and (ii) an online protection phase that leverages the learned PPT to obfuscate images before they are shared online.

\noindent\textbf{Offline Learning.} \scheme{} learns a unique 3D facial signature for a user $\mathcal{U}$ using a small set of facial images $\boldsymbol{\Omega}$. This signature is encapsulated into a 2D representation called the privacy protection texture (PPT), denoted by $\boldsymbol{\mathcal{T}}$. The PPT is designed to be reusable across different images of the same user, including those not seen during training. This generalization is achieved by iteratively fine-tuning $\boldsymbol{\mathcal{T}}$ to provide consistent protection across training images against an ensemble of FR models $\boldsymbol{\mathcal{F}}$. In particular, at the $t$-th iteration, \scheme{} samples a mini-batch $\boldsymbol{\mathcal{B}}\subset\boldsymbol{\Omega}$ and updates the PPT:
\begin{equation}\label{eq:train}
	\boldsymbol{\mathcal{T}}^{t+1}=\underset{[-\epsilon, \epsilon]}{\textsc{Clip}}\big[\boldsymbol{\mathcal{T}}^t-\eta\textsc{Sign}(\nabla_{\boldsymbol{\mathcal{T}}^t}\mathcal{L}(\boldsymbol{\mathcal{B}}, \boldsymbol{\mathcal{T}}^t; \boldsymbol{\mathcal{F}})\big],
\end{equation}
where $\epsilon$ is the $L_\infty$ perturbation bound, $\eta$ is the learning rate, and $\mathcal{L}$ is the \scheme{} loss function introduced in Section~\ref{sec:prot}.

\noindent\textbf{Online Protection.} The learned PPT $\boldsymbol{\mathcal{T}}$ can be applied to protect any future facial image of user $\mathcal{U}$. Given a photo to be protected, a lightweight face detector (e.g., MediaPipe~\cite{lugaresi2019mediapipe}) first locates the face region $\boldsymbol{x}$. \scheme{} then deforms the PPT to align with the pose and expression of $\boldsymbol{x}$ before applying it to generate the protected face:
\begin{equation}\label{eq:online}
\Theta(	\boldsymbol{x};\boldsymbol{\mathcal{T}})=\underset{[0,1]}{\textsc{Clip}}\big[\boldsymbol{x}-\delta(\boldsymbol{\mathcal{T}}; \boldsymbol{x})\big],
\end{equation}
where $\delta$ is the deformation function described in Section~\ref{sec:vis}.

\subsection{\scheme{} Optimization Loss}\label{sec:prot}
\scheme{} guides the learning of PPT using the loss function $\mathcal{L}$ (Equation~\ref{eq:train}) composed of two optimization components.

\begin{figure}\centering
	\includegraphics[width=0.495\linewidth]{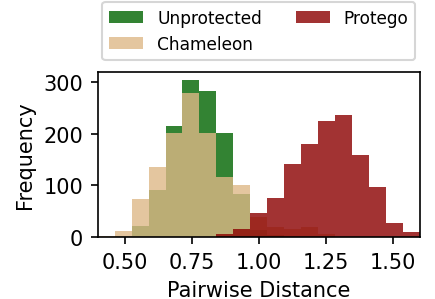} 
	\includegraphics[width=0.495\linewidth]{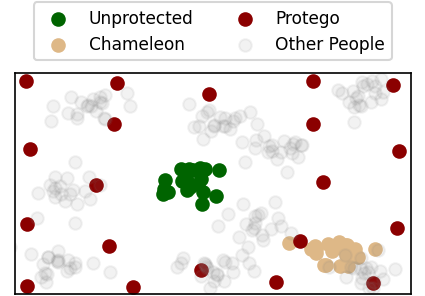} 
	\caption{Chameleon-protected images have highly similar features, as reflected by their small pairwise distances comparable to those of unprotected images. It results in dense clusters in the feature space (see the scatter plot). In contrast, \scheme{}-protected images exhibit diverse features, preventing protected queries from retrieving protected DB entries.}\label{fig:prot-opt}
\end{figure}

\noindent\textbf{Protection Optimization.} Prior approaches optimize their perturbations to make protected and unprotected images dissimilar in the feature space. While this objective is necessary to ensure that an unprotected query cannot retrieve protected DB entries (and vice versa), it has a critical shortcoming: protected images tend to form a dense cluster in the feature space (Figure~\ref{fig:prot-opt}). As a result, a protected query can still retrieve other protected entries with high accuracy.
To overcome this, \scheme{} generates perturbations that deceive the FR model into being hypersensitive to variations in protected content. Small changes in input (e.g., micro expressions of the same individual) should yield significantly different features. This can be achieved by diversifying the feature vectors of protected images.
Let $\{F(\Theta(\boldsymbol{x}_i;\boldsymbol{\mathcal{T}}^t)) \mid \boldsymbol{x}_i \in \boldsymbol{\mathcal{B}}\}$ denote the set of feature vectors extracted from protected images by the FR model $F$. The corresponding Gram matrix $\boldsymbol{G}(\boldsymbol{\mathcal{B}}, \boldsymbol{\mathcal{T}}^t; F)$ is defined by $G_{i,j}=F(\Theta(\boldsymbol{x}_i;\boldsymbol{\mathcal{T}}^t))^{\intercal}F(\Theta(\boldsymbol{x}_j;\boldsymbol{\mathcal{T}}^t))$. Since the determinant of the Gram matrix reflects the volume spanned by the feature vectors, \scheme{}'s protection loss is defined as:
\begin{equation}
	\begin{split}
		&\mathcal{L}_{\text{Protect}}(\boldsymbol{\mathcal{B}}, \mathcal{\boldsymbol{\mathcal{T}}}^t;\boldsymbol{\mathcal{F}})=\frac{-1}{\vert\vert\boldsymbol{\mathcal{F}}\vert\vert}\sum_{F\in\boldsymbol{\mathcal{F}}}\log\det\boldsymbol{G}(\boldsymbol{\mathcal{B}},\boldsymbol{\mathcal{T}}^t;F) \\
		&+ \frac{1}{\vert\vert\boldsymbol{\mathcal{F}}\vert\vert\vert\vert\boldsymbol{\mathcal{B}}\vert\vert}\sum_{F\in\boldsymbol{\mathcal{F}}}\sum_{\boldsymbol{x}\in\boldsymbol{\mathcal{B}}}\textsc{Sim}(F(\boldsymbol{x}), F(\Theta(\boldsymbol{x};\boldsymbol{\mathcal{T}}^t))).
	\end{split}
\end{equation}
The 1st term promotes orthonormality among protected features by maximizing their volume, and the 2nd term penalizes similarity between protected and unprotected images.

\noindent\textbf{Perceptual Optimization.} To ensure that the generated PPT minimally disrupts visual appearance, we constrain the structural similarity (SSIM~\cite{wang2004image}) between protected and unprotected images. Specifically, we control the average SSIM degradation to stay within a threshold. This is formalized by the following perceptual loss:
\begin{equation}
	\begin{split}
		&\mathcal{L}_{\text{Percept}}(\boldsymbol{\mathcal{B}}, \boldsymbol{\mathcal{T}}^t)\\
		&=\max\bigg[\sum_{\boldsymbol{x}\in\boldsymbol{\mathcal{B}}}\frac{1-\textsc{SSIM}(\boldsymbol{x}, \Theta(\boldsymbol{x}; \boldsymbol{\mathcal{T}}^t))}{2\vert\vert\boldsymbol{\mathcal{B}}\vert\vert}-\omega, 0\bigg],
	\end{split}
\end{equation}
where $\omega$ is a user-defined threshold controlling the maximum allowable drop in SSIM.

\noindent\textbf{Overall.} The \scheme{} loss combines two components:
\begin{equation}\small
	\begin{split}
		\mathcal{L}(\boldsymbol{\mathcal{B}}, \boldsymbol{\mathcal{T}}^t;\boldsymbol{\mathcal{F}})=\mathcal{L}_\text{Protect}(\boldsymbol{\mathcal{B}}, \mathcal{\boldsymbol{\mathcal{T}}}^t;\boldsymbol{\mathcal{F}}) + 
		\lambda_{\text{SSIM}}\mathcal{L}_\text{Percept}(\boldsymbol{\mathcal{B}}, \boldsymbol{\mathcal{T}}^t),
	\end{split}
\end{equation}
where $\lambda_{\text{SSIM}}$ balances protection effectiveness and perceptual quality, and we use dynamic scheduling~\cite{shan2020fawkes} such that it will be adjusted automatically.

\subsection{3D Facial Privacy Protection}\label{sec:vis}
Existing methods assume the face to be front-facing and are most suitable for protecting such images. However, when faces exhibit varying poses or expressions, the perturbations do not adapt accordingly and may appear as phantom faces, as shown in the demo videos to be discussed in Section~\ref{sec:eval-natural}. This lack of adaptability compromises visual coherence, an issue that becomes particularly pronounced in videos.

To overcome these limitations, we leverage the concept of a UV map~\cite{foley1996computer}, a standardized 2D coordinate system used to represent the surface of a 3D face in a flattened form. Each location in this space corresponds to a specific semantic region of the human face (e.g., the left eye, nose tip, or jawline)~\cite{ding2016comprehensive,deng2018uv,li2017learning}. In our context, UV mapping serves as a canonical facial layout that enables us to learn how to perturb semantic regions in a pose-invariant way. By training on images captured from diverse angles, \scheme{} learns how to apply identity-aware, region-specific perturbations (e.g., how to modify the left ear) that can later be projected back to faces with varying poses, maintaining visual coherence.

\begin{figure}\centering
	\includegraphics[width=\linewidth]{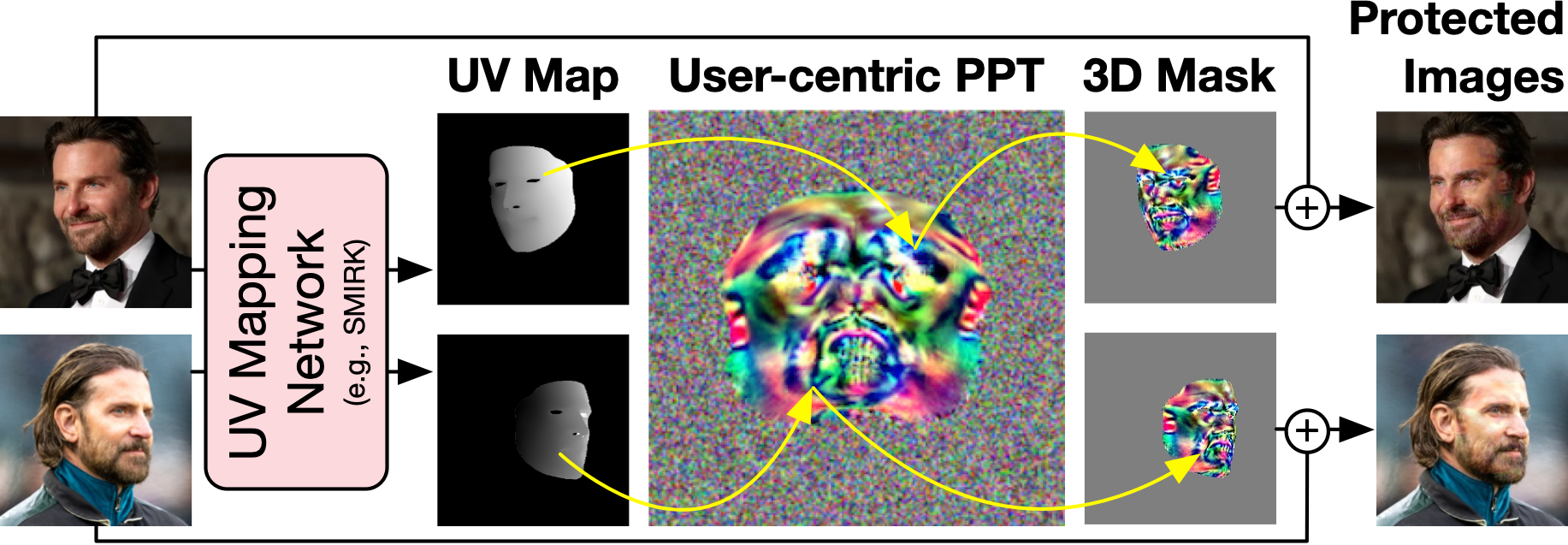} 
\caption{\scheme{}’s PPT is pose-invariant. Given an image to be protected, we extract its UV map using a pretrained UV mapping network. This UV map guides the sampling of the PPT to generate a perturbation aligned with the image’s pose and expression.}\label{fig:vis-pro}
\end{figure}

As shown in Figure~\ref{fig:vis-pro}, given a facial image $\boldsymbol{x}$ to be protected, we employ a pretrained UV mapping network (e.g., SMIRK~\cite{retsinas2024smirk} used in this paper) to estimate its UV map $\boldsymbol{UV}(\boldsymbol{x})$, which projects each pixel in $\boldsymbol{x}$ to a canonical UV space where \scheme{}'s pose-agnostic PPT $\boldsymbol{\mathcal{T}}$ is defined. To adapt $\boldsymbol{\mathcal{T}}$ to the specific pose and expression of the face in $\boldsymbol{x}$ (see Equation~\ref{eq:online}), we apply a grid sampling operation to generate the 3D mask $\delta(\boldsymbol{\mathcal{T}};\boldsymbol{x})= \textsc{GridSample}(\boldsymbol{\mathcal{T}}, \boldsymbol{UV}(\boldsymbol{x}))$, which has the same spatial dimensions as the input image $\boldsymbol{x}$ and specifies how the perturbation should be applied in the image space. This operation is differentiable. Hence, during the training of PPT, the gradients can flow to the UV space and construct the pose-invariant protection texture. 

\begin{table}[t]\centering\small
	\begin{tabular}{|lcc|c|}
		\hline
		\multicolumn{1}{|c|}{\textbf{Name}} & \multicolumn{1}{c|}{\textbf{\begin{tabular}[c]{@{}c@{}}Query\\ Images\end{tabular}}} & \textbf{\begin{tabular}[c]{@{}c@{}}Training \&\\ DB Entries\end{tabular}} & \textbf{\begin{tabular}[c]{@{}c@{}}DB\\ Entries\end{tabular}} \\ \hline
		\multicolumn{1}{|l|}{Bradley Cooper}              & \multicolumn{1}{c|}{23}                                                                   &           69                                                      &    23                                                          \\ \hline
		\multicolumn{1}{|l|}{Hugh Grant}              & \multicolumn{1}{c|}{22}                                                                   &           67                                                      &        23                                                      \\ \hline
		\multicolumn{1}{|l|}{Debra Messing}              & \multicolumn{1}{c|}{26}                                                                   &            78                                                     &     26                                                         \\ \hline
		\multicolumn{1}{|l|}{Felicity Huffman}              & \multicolumn{1}{c|}{27}                                                                   &     81                                                            &       27                                                       \\ \hline
		\multicolumn{4}{|c|}{... 16 more users ...}                                                   \\ \hline
		\multicolumn{1}{|l|}{Others}        & \multicolumn{1}{c|}{/}                                                                  & /                                                               &    40679                                                          \\ \hline
		\multicolumn{3}{|r|}{\textbf{Total Number of DB Entries:}}                                                                                                                                                &                               41176                               \\ \hline
	\end{tabular}
	\caption{Four example users (celebrities) on FaceScrub.}\label{tab:fs-users}
\end{table}
\begin{table}[t]\small\centering
\setlength\tabcolsep{1.5pt}
\begin{tabular}{|c|c|c|c|cc|}
	\hline
	\multirow{2}{*}{\textbf{ID}} & \multirow{2}{*}{\textbf{\begin{tabular}[c]{@{}c@{}}FR\\ Method\end{tabular}}} & \multirow{2}{*}{\textbf{\begin{tabular}[c]{@{}c@{}}Neural\\ Arch.\end{tabular}}} & \multirow{2}{*}{\textbf{\begin{tabular}[c]{@{}c@{}}Training\\ Dataset\end{tabular}}} & \multicolumn{2}{c|}{\textbf{Recall (\%)}}              \\ \cline{5-6} 
	&                                                                              &                                                                                  &                                                                                      & \multicolumn{1}{c|}{\textbf{\tiny FaceScrub}} & \textbf{\tiny LFW} \\ \hline
	AD-IR18-4M & AdaFace                                                                      & IR18                                                                             & WebFace4M                                                                            & \multicolumn{1}{c|}{74.77}                   &          81.55    \\ \hline
	AD-IR50-4M & AdaFace                                                                      & IR50                                                                             & WebFace4M                                                                            & \multicolumn{1}{c|}{93.89}                   &   95.45           \\ \hline
	AD-IR50-CA & AdaFace                                                                      & IR50                                                                             & CAS-WebFace                                                                        & \multicolumn{1}{c|}{71.68}                   &           70.09   \\ \hline
	AD-IR50-MS & AdaFace                                                                      & IR50                                                                             & MS1MV2                                                                               & \multicolumn{1}{c|}{94.68}                   &           88.45   \\ \hline
	AD-IR100-4M & AdaFace                                                                      & IR100                                                                            & WebFace4M                                                                            & \multicolumn{1}{c|}{97.36}                   & 97.22             \\ \hline
	AR-IR50-CA & ArcFace                                                                      & IR50                                                                             & CAS-WebFace                                                                        & \multicolumn{1}{c|}{51.03}                   &   44.12           \\ \hline
	AR-MN-CA & ArcFace                                                                      & MNet                                                                        & CAS-WebFace                                                                        & \multicolumn{1}{c|}{71.70}                   &    78.11          \\ \hline
	AR-MFN-CA & ArcFace                                                                      & MFNet                                                                    & CAS-WebFace                                                                        & \multicolumn{1}{c|}{49.13}                   &    54.78          \\ \hline
	CS-IR50-CA & CosFace                                                                      & IR50                                                                             & CAS-WebFace                                                                        & \multicolumn{1}{c|}{79.51}                   & 79.21              \\ \hline
	SM-IR50-CA & Softmax                                                                      & IR50                                                                             & CAS-WebFace                                                                        & \multicolumn{1}{c|}{64.10}                   &   70.57           \\ \hline
\end{tabular}
\caption{The collection of publicly available pretrained FR models for demonstrating \scheme{}'s effectiveness.}\label{tab:fr-collection}
\end{table}
\section{Evaluation}
\noindent\textbf{Datasets.} We conduct experiments on two benchmark datasets having facial images of celebrities: FaceScrub~\cite{ng2014data} and LFW~\cite{huang2008labeled}. Following the evaluation strategy adopted by \citet{chow2024personalized}, we randomly select 20 celebrities as \scheme{} users. Table~\ref{tab:fs-users} lists a subset of them on FaceScrub as examples. For each user, their facial images are divided into three parts: (i) 20\% are used as queries by the privacy intruder, (ii) 60\% are DB entries used to train the PPT in \scheme{}, and (iii) the remaining 20\% are also DB entries but remain unseen during the PPT training. This last portion is critical for evaluating the protection of unseen images. All other individuals in the dataset are included in the database as noise.

\noindent\textbf{FR Models.} Privacy intruders typically deploy pretrained FR models for two reasons: (i) public cloud services like Azure now require manual approval to ensure ethical use of their FR services, and (ii) many high-quality pretrained models are readily available and free to use. Therefore, our experiments focus on the collection of publicly available pretrained FR models listed in Table~\ref{tab:fr-collection}. This list is carefully curated to demonstrate \scheme{}'s effectiveness across a variety of FR models, covering different FR methods (loss functions), neural architectures, and training datasets. 

\noindent\textbf{Metrics.} We define a customized ``recall" metric to evaluate the performance of an FR-driven retrieval system~\cite{han2022data}. Specifically, for a person with $K$ DB entries, we issue a query to the system, retrieve the top-$K$ most similar entries, and compute the percentage of relevant retrievals. Averaging this value across all queries indicates how effectively the system retrieves entries associated with that person. \scheme{} is designed to weaken this capability for its users. Table~\ref{tab:fr-collection} (5th and 6th columns) reports the average recall across 20 users for each dataset, serving as the baseline performance in the absence of any privacy protection.

\noindent\textbf{Default Setting.} We adopt a black-box protection setting in which \scheme{} has no access to the FR model used by the privacy intruder's retrieval system. The default FR model used by the intruder is AD-IR50-CA, while the remaining models in Table~\ref{tab:fr-collection} are used to train the PPT in \scheme{} with $\epsilon=0.063$, $\eta=\epsilon/10$, $\omega=0.025$, $\vert\vert\boldsymbol{\mathcal{B}}\vert\vert=4$. In Section~\ref{sec:eval-transfer}, we will show that \scheme{} can protect against a variety of unknown FR models with only a small team for training. 

The source code of \scheme{} is available at \url{https://github.com/HKU-TASR/Protego}.

\subsection{Effective and Consistent Protection}
\begin{figure}[t]
	\centering
	\begin{subfigure}{\linewidth}
		\centering
		\includegraphics[width=0.85\linewidth]{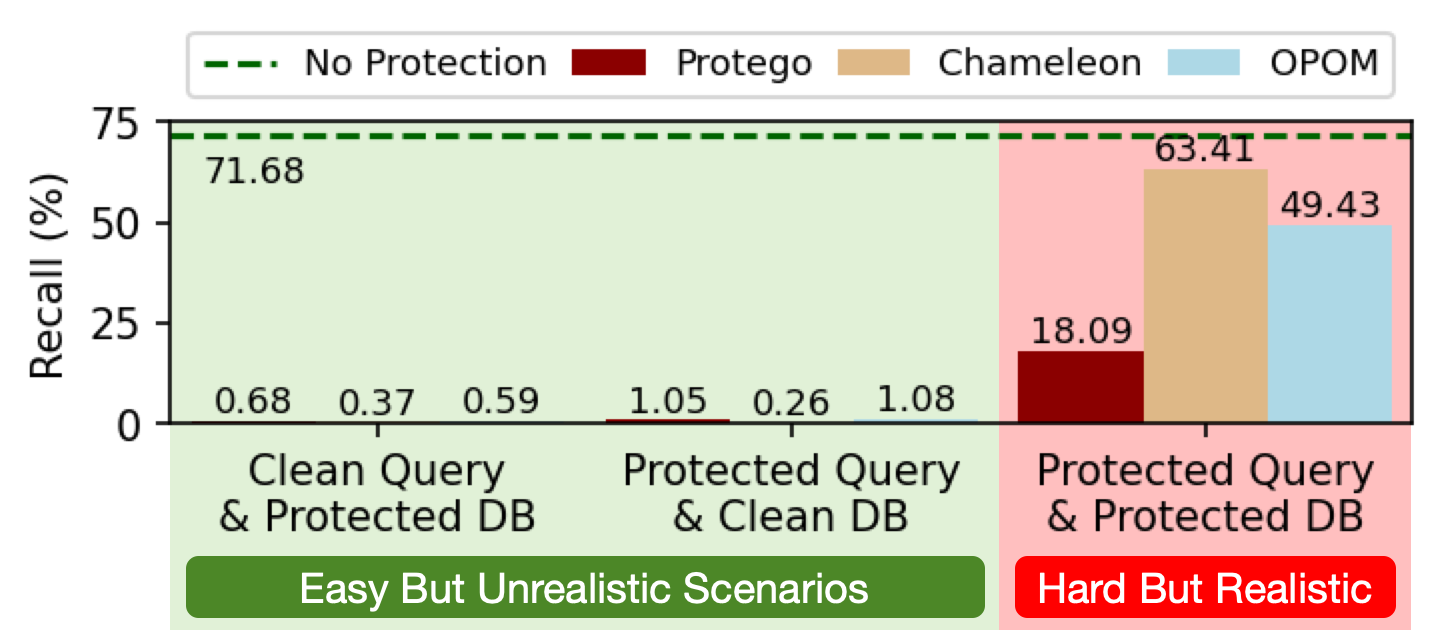}
		\caption{Dataset: FaceScrub}\label{fig:psr-fs}
	\end{subfigure}
	\begin{subfigure}{\linewidth}
		\centering
		\includegraphics[width=0.85\linewidth]{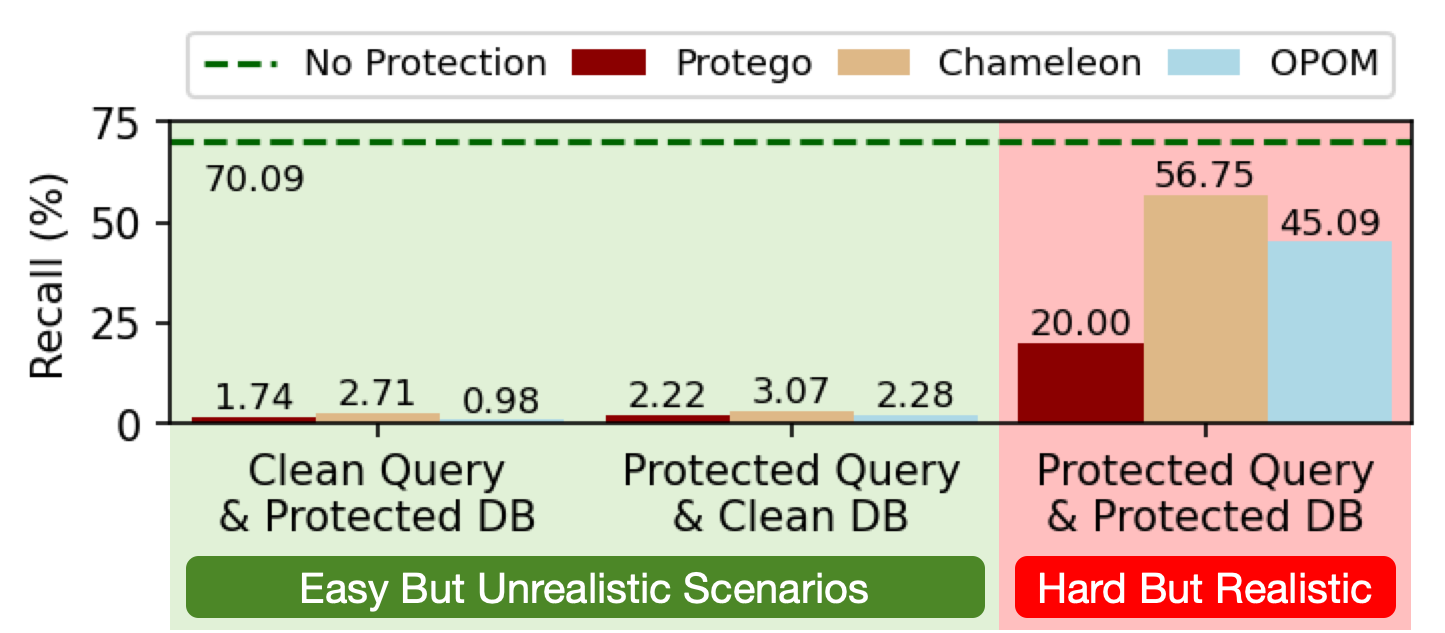}
		\caption{Dataset: LFW}\label{fig:psr-lfw}
	\end{subfigure}%
	\caption{\scheme{} offers effective protection on both datasets across three scenarios: (i) unprotected queries with protected DB, (ii) protected queries with unprotected DB, and (iii) protected queries with protected DB. Existing methods perform well under easy but unrealistic scenarios, but they fail to prevent protected queries from matching protected DB entries.}\label{fig:psr-all}
\end{figure}
\begin{figure}[t]
\centering
\includegraphics[width=0.905\linewidth]{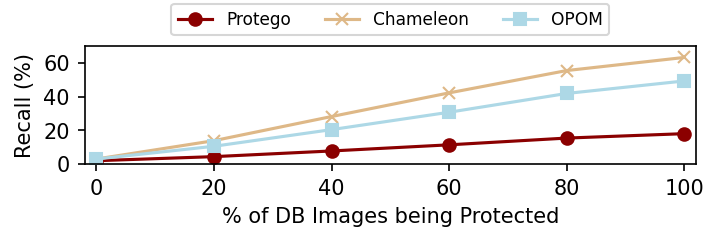}
\caption{As more DB entries are protected, Protego continues to ensure low recall, as protected queries do not retrieve protected entries, a limitation common in other methods.}\label{fig:leaked-portion}
\end{figure}
\begin{figure}[t]
\centering
\includegraphics[width=0.9\linewidth]{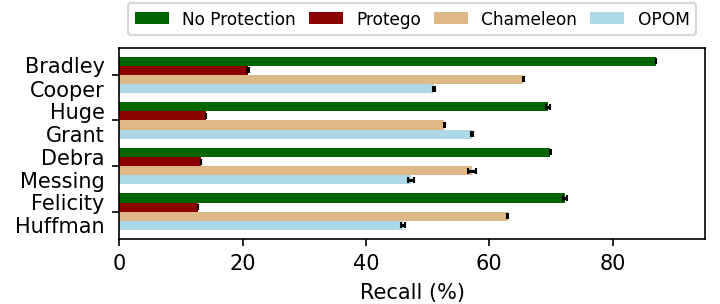}
\caption{\scheme{} provides effective and consistent protection across users. The error bars indicate the standard deviation over five runs with different random seeds.}\label{fig:user-psr}
\end{figure}
\begin{table}[t]\centering\small
\setlength\tabcolsep{1.4pt} 
\begin{tabular}{|c|ccccc|}
	\hline
	\multirow{2}{*}{\textbf{Query}}                        & \multicolumn{5}{c|}{\textbf{Retrieved DB Entries}}                                                                                                                                                                                                                                                                                                                                         \\ \cline{2-6} 
	& \multicolumn{1}{c|}{\textbf{Rank 1}}                                         & \multicolumn{1}{c|}{\textbf{Rank 2}}                                         & \multicolumn{1}{c|}{\textbf{Rank 3}}                                         & \multicolumn{1}{c|}{\textbf{Rank 4}}                                         & \textbf{Rank 5}                                         \\ \hline
	\multicolumn{6}{|l|}{(a) Protego} \\ \cline{1-6} 
	\begin{tabular}[c]{@{}c@{}}\includegraphics[width=34pt]{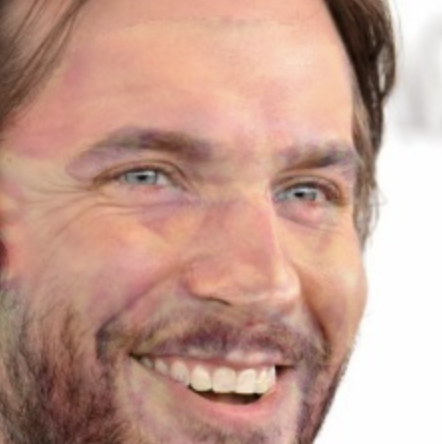}\\ {\scriptsize B. Cooper}\end{tabular} & \multicolumn{1}{c|}{\begin{tabular}[c]{@{}c@{}}\includegraphics[width=34pt]{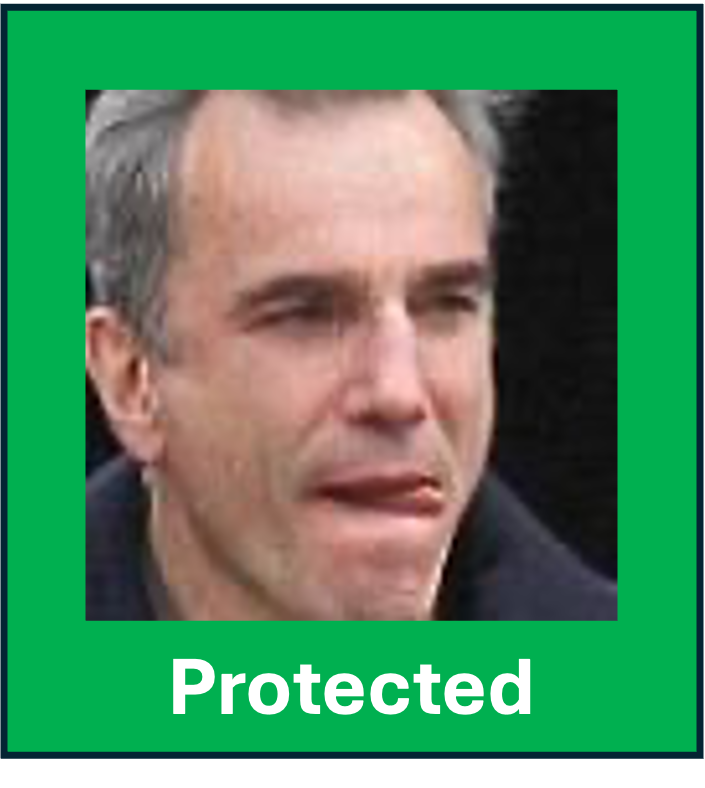}\\ {\scriptsize Day-Lewise}\end{tabular}} & \multicolumn{1}{c|}{\begin{tabular}[c]{@{}c@{}}\includegraphics[width=34pt]{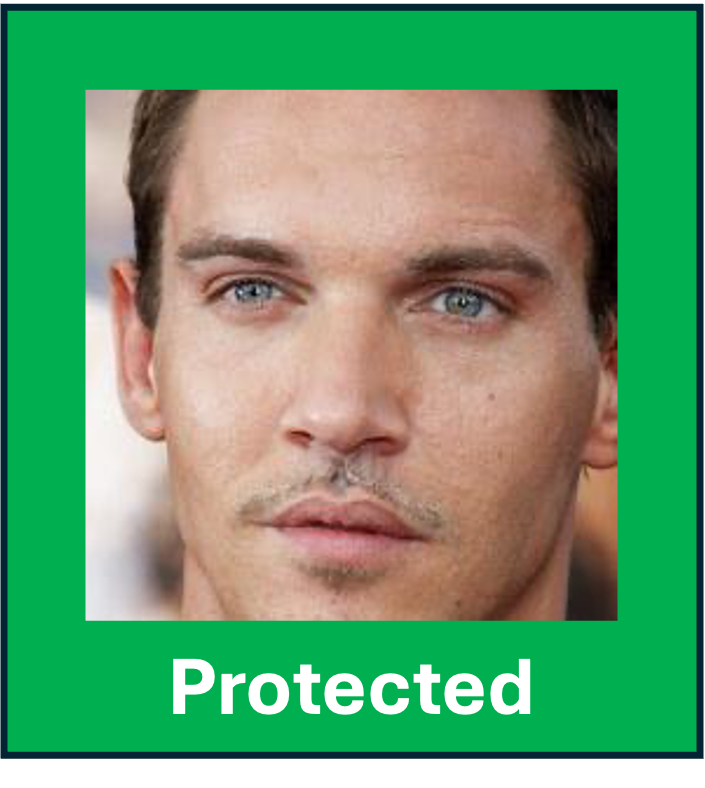}\\ {\scriptsize J. Meyers}\end{tabular}} & \multicolumn{1}{c|}{\begin{tabular}[c]{@{}c@{}}\includegraphics[width=34pt]{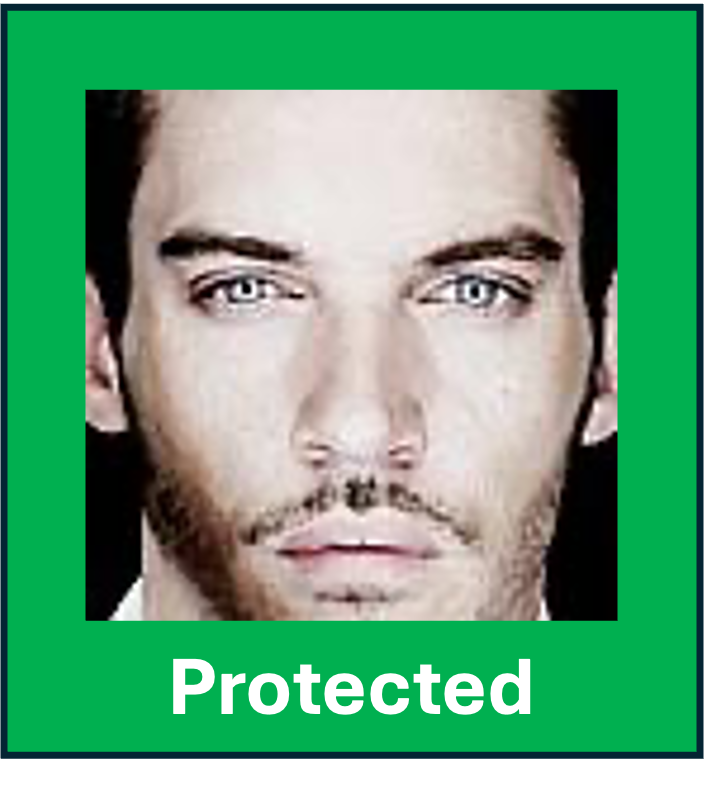}\\ {\scriptsize J. Meyers}\end{tabular}} & \multicolumn{1}{c|}{\begin{tabular}[c]{@{}c@{}}\includegraphics[width=34pt]{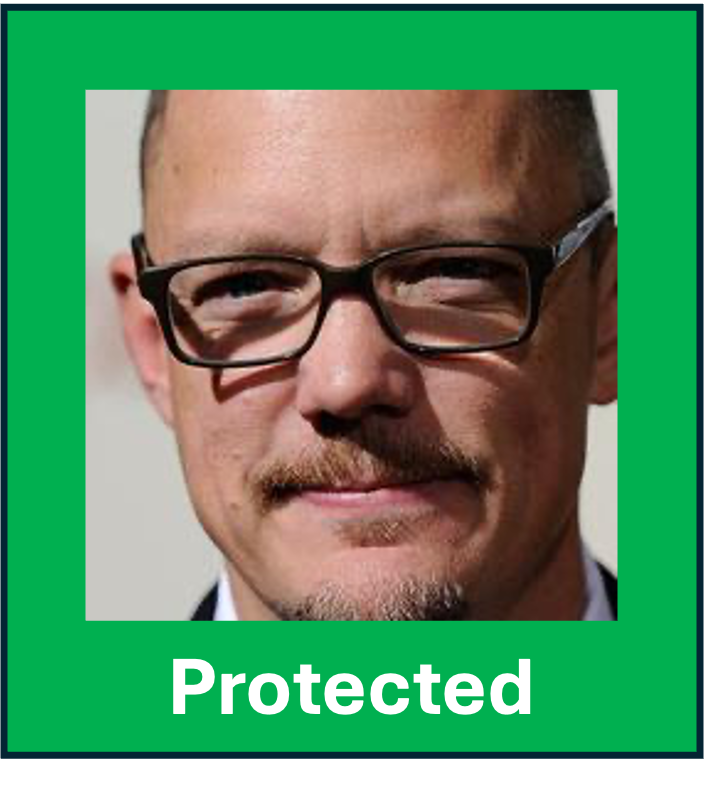}\\ {\scriptsize M. Lillarde}\end{tabular}} & \begin{tabular}[c]{@{}c@{}}\includegraphics[width=34pt]{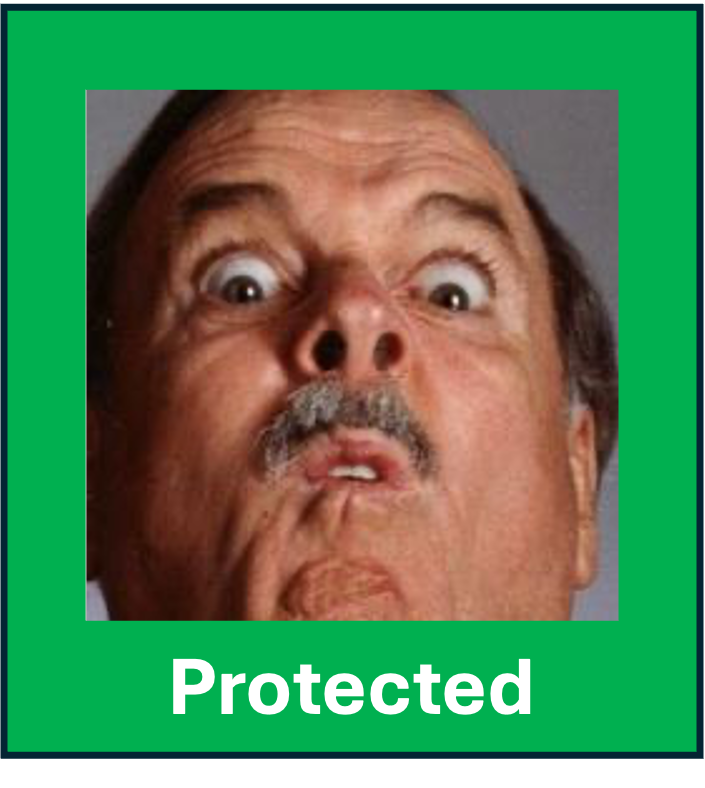}\\ {\scriptsize J. Cleese}\end{tabular} \\ \cline{1-6} 
	
	\multicolumn{6}{|l|}{(b) Chameleon} \\ \cline{1-6} 
	\begin{tabular}[c]{@{}c@{}}\includegraphics[width=34pt]{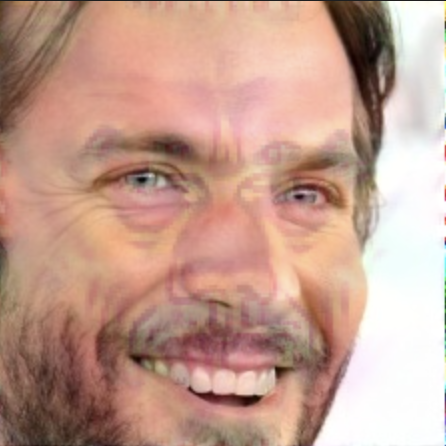}\\ {\scriptsize B. Cooper}\end{tabular}  & \multicolumn{1}{c|}{\begin{tabular}[c]{@{}c@{}}\includegraphics[width=34pt]{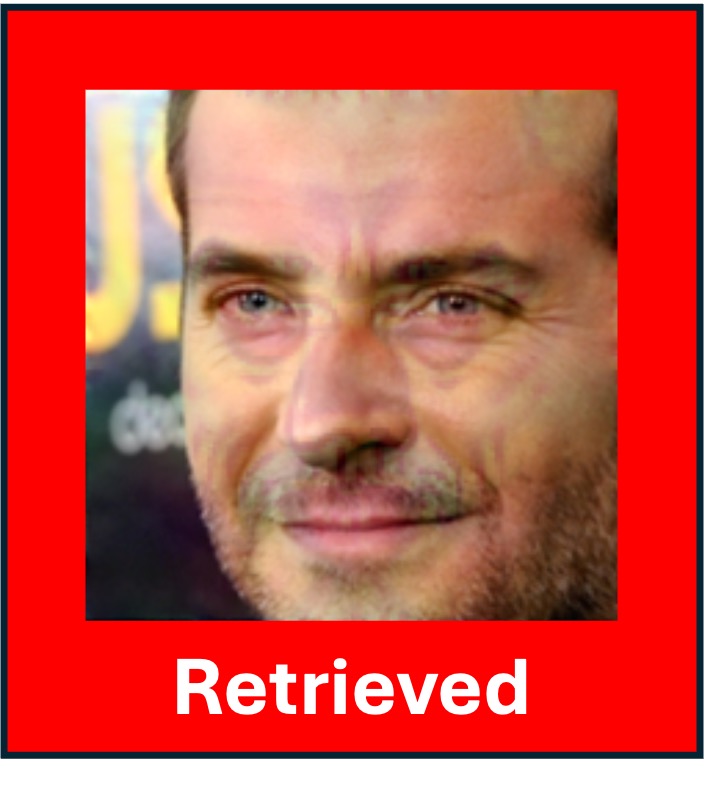}\\ {\scriptsize B. Cooper}\end{tabular}} & \multicolumn{1}{c|}{\begin{tabular}[c]{@{}c@{}}\includegraphics[width=34pt]{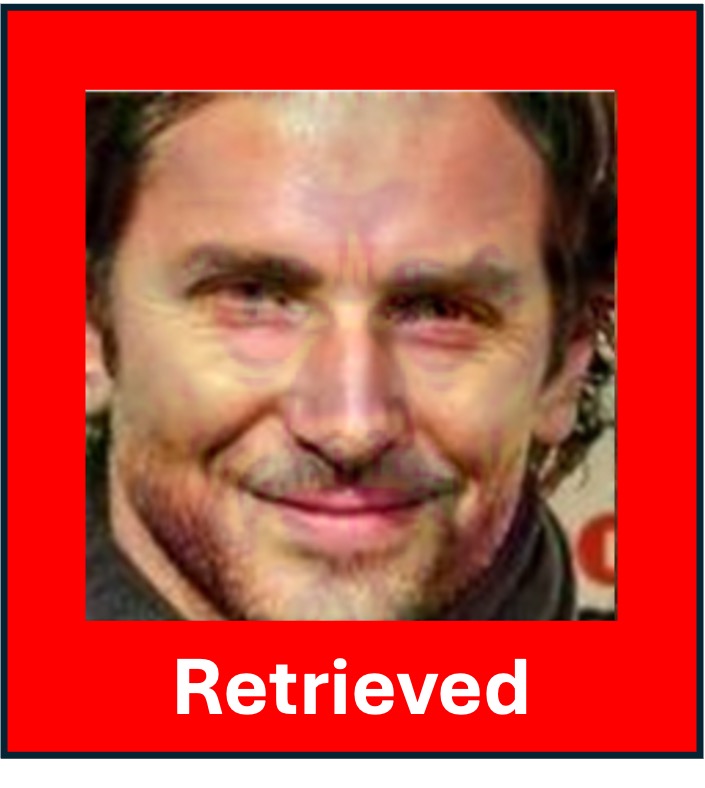}\\ {\scriptsize B. Cooper}\end{tabular}} & \multicolumn{1}{c|}{\begin{tabular}[c]{@{}c@{}}\includegraphics[width=34pt]{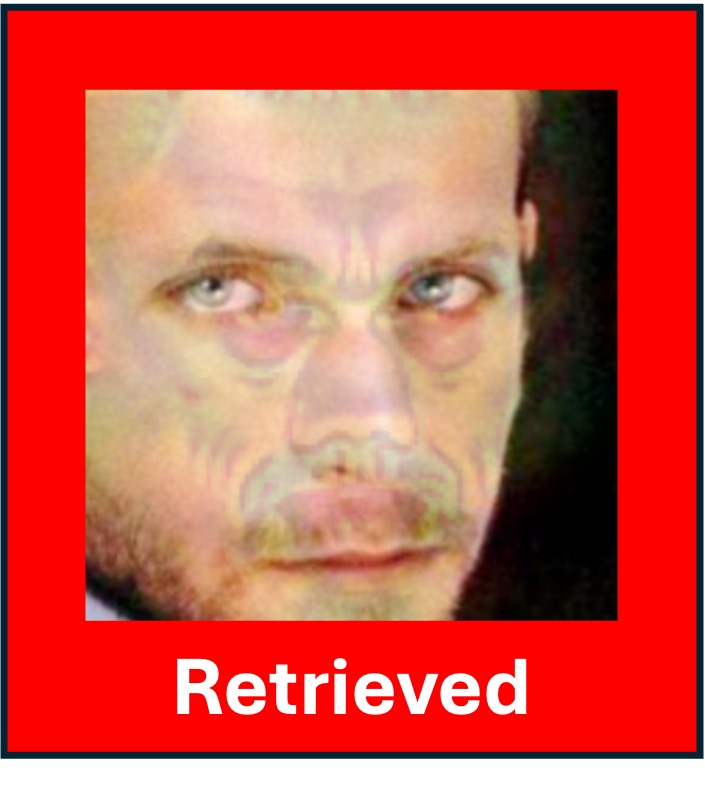}\\ {\scriptsize B. Cooper}\end{tabular}} & \multicolumn{1}{c|}{\begin{tabular}[c]{@{}c@{}}\includegraphics[width=34pt]{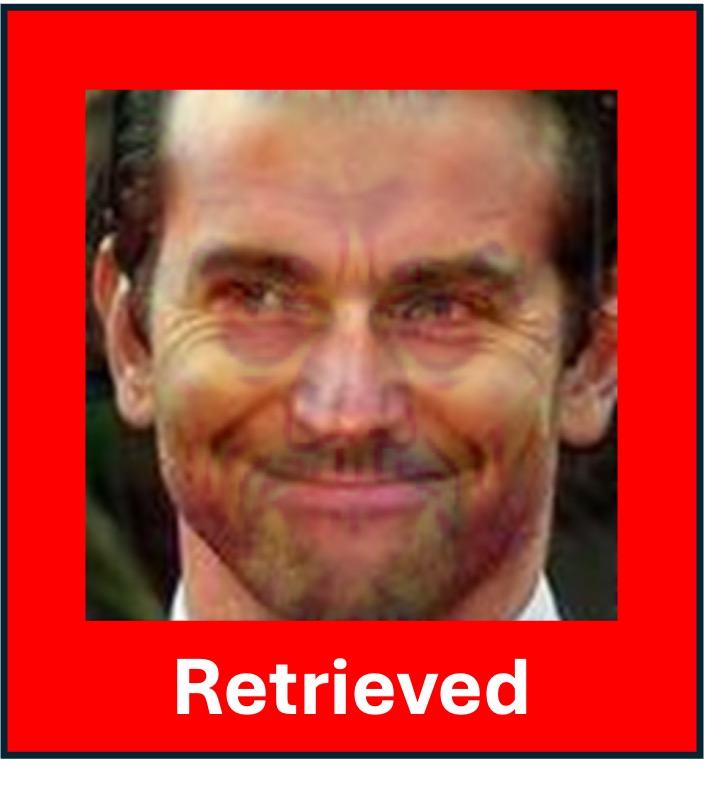}\\ {\scriptsize B. Cooper}\end{tabular}} & \begin{tabular}[c]{@{}c@{}}\includegraphics[width=34pt]{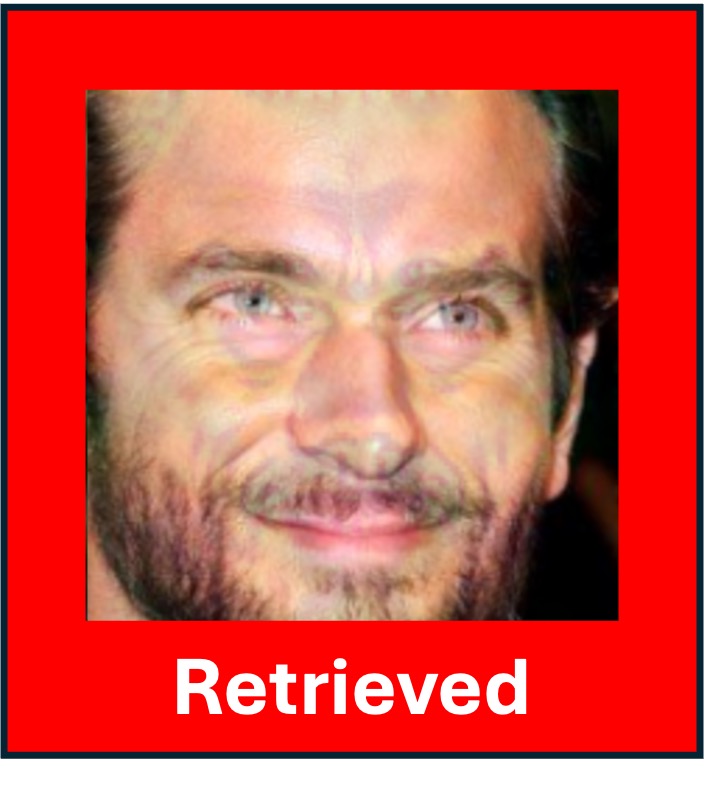}\\ {\scriptsize B. Cooper}\end{tabular} \\ \cline{1-6} 
	\multicolumn{6}{|l|}{(c) OPOM} \\ \cline{1-6} 
	\begin{tabular}[c]{@{}c@{}}\includegraphics[width=34pt]{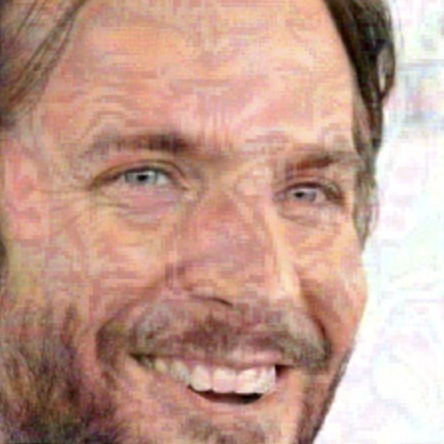}\\ {\scriptsize B. Cooper}\end{tabular} & \multicolumn{1}{c|}{\begin{tabular}[c]{@{}c@{}}\includegraphics[width=34pt]{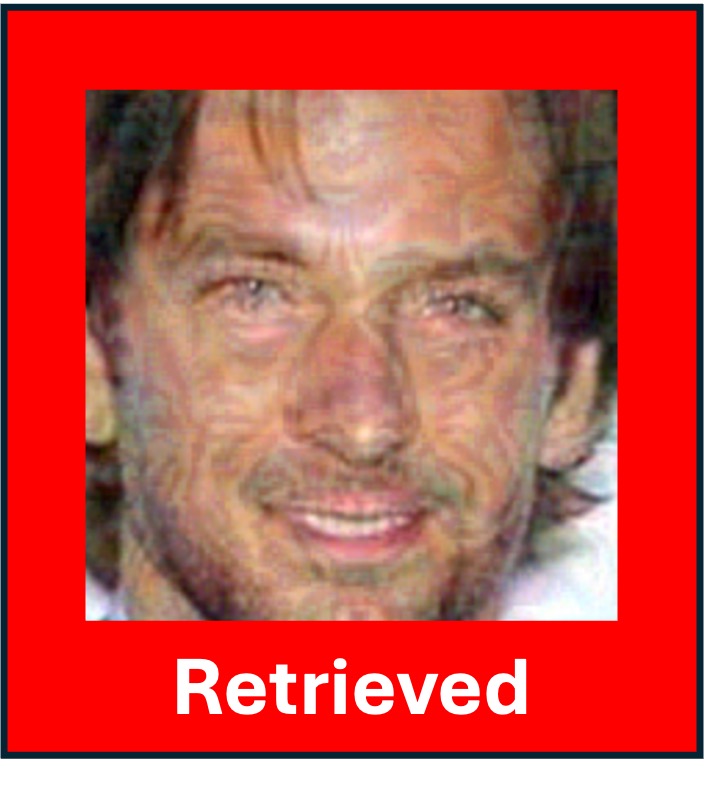}\\ {\scriptsize B. Cooper}\end{tabular}} & \multicolumn{1}{c|}{\begin{tabular}[c]{@{}c@{}}\includegraphics[width=34pt]{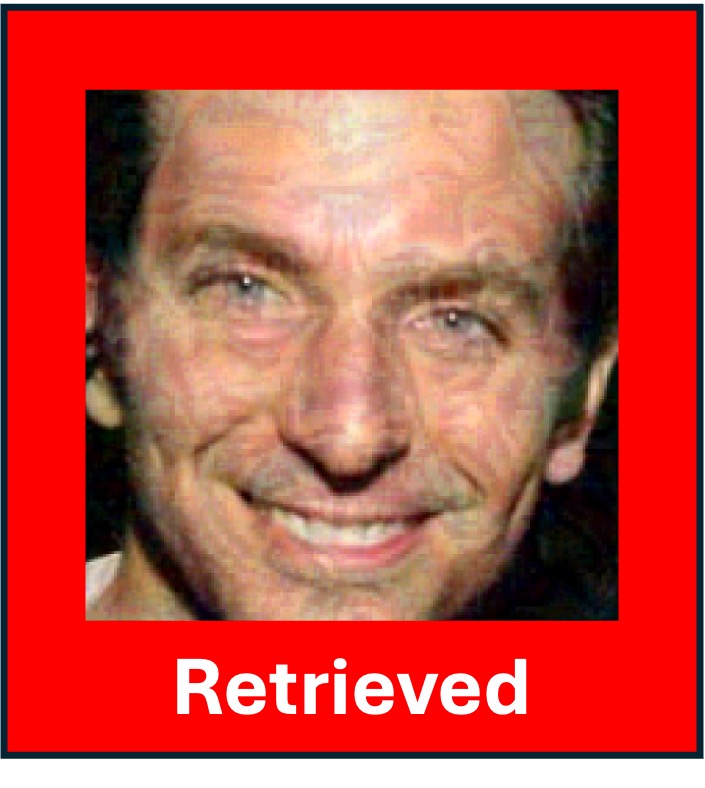}\\ {\scriptsize B. Cooper}\end{tabular}} & \multicolumn{1}{c|}{\begin{tabular}[c]{@{}c@{}}\includegraphics[width=34pt]{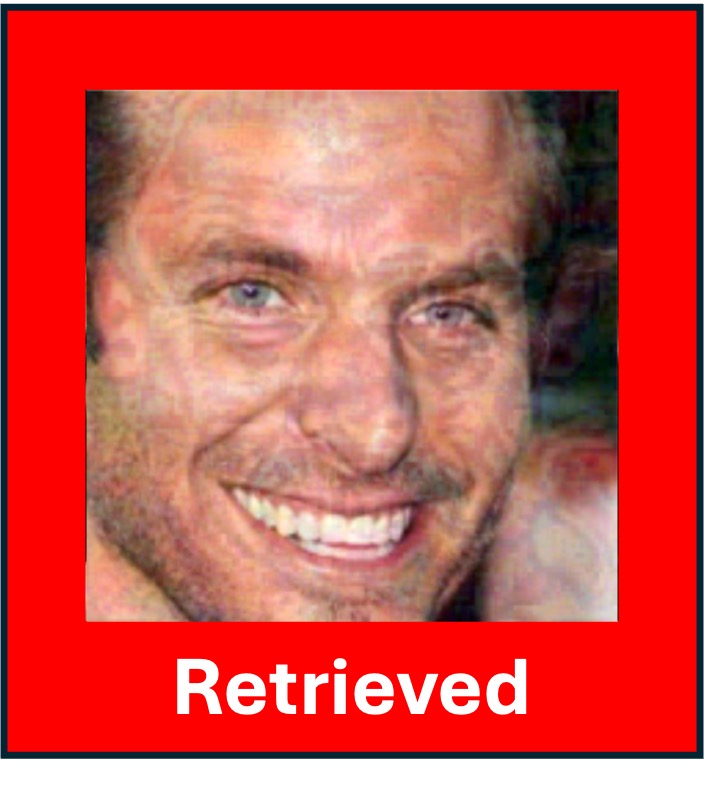}\\ {\scriptsize B. Cooper}\end{tabular}} & \multicolumn{1}{c|}{\begin{tabular}[c]{@{}c@{}}\includegraphics[width=34pt]{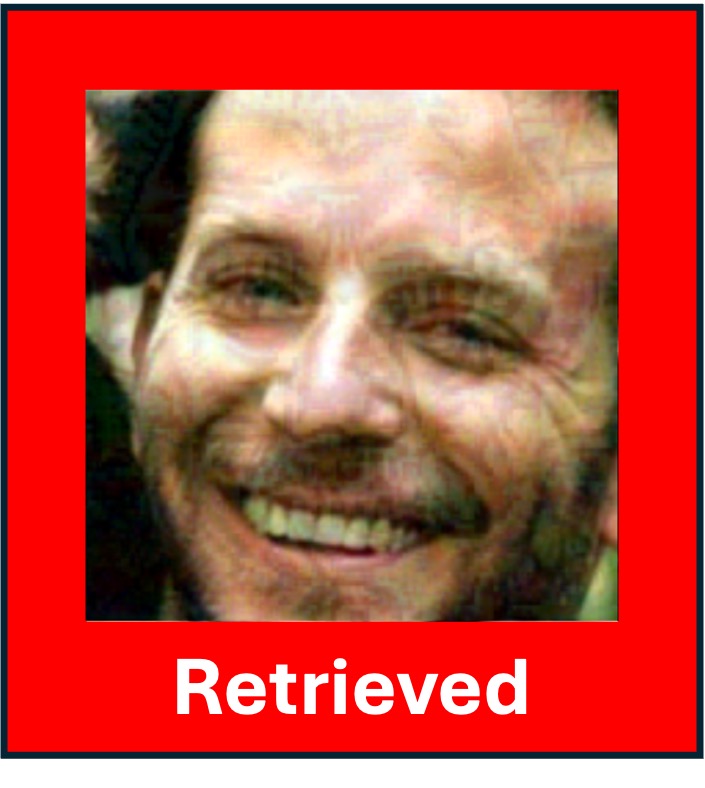}\\ {\scriptsize B. Cooper}\end{tabular}} & \begin{tabular}[c]{@{}c@{}}\includegraphics[width=34pt]{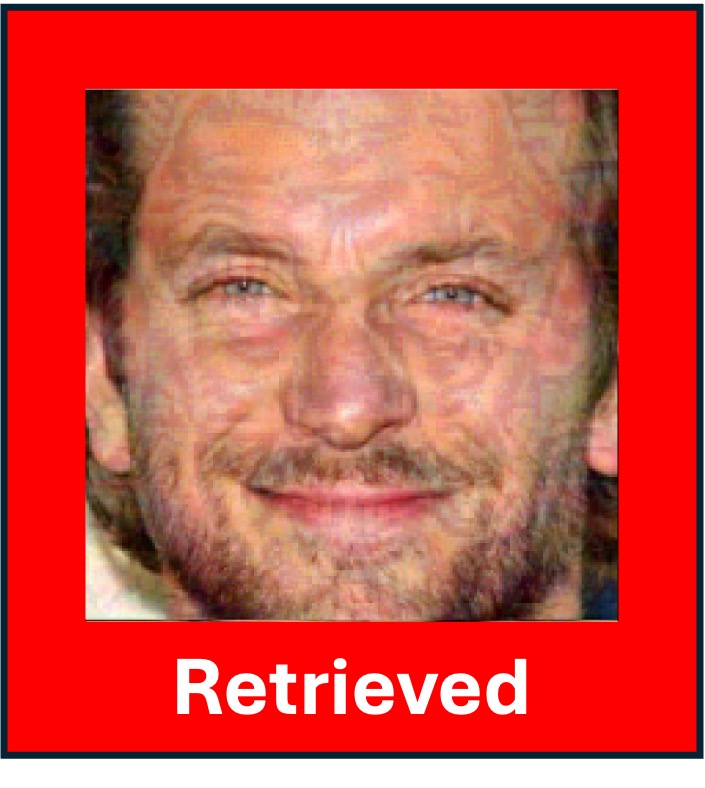}\\ {\scriptsize B. Cooper}\end{tabular} \\ \hline
\end{tabular}
\caption{The \scheme{}-protected query (a) does not match any DB entry of the same person, whether protected or not. In contrast, existing methods such as Chameleon (b) and OPOM (c) still allow retrieval of protected DB entries belonging to the same individual.}\label{tab:matching-baselines}
\end{table}

\begin{table*}[t]\centering\small
	\setlength\tabcolsep{1.5pt}
	\begin{tabular}{|@{}p{45pt}@{}|c|ccc|ccc|ccc|}
		\hline
		\multicolumn{1}{|c|}{\multirow{2}{*}{\textbf{User}}} & \multirow{2}{*}{\textbf{\begin{tabular}[c]{@{}c@{}}\scheme{}\\PPT\end{tabular}}} & \multicolumn{3}{c|}{\textbf{Sample Image 1}}                                                             & \multicolumn{3}{c|}{\textbf{Sample Image 2}}                                                             & \multicolumn{3}{c|}{\textbf{Sample Image 3}}                                                             \\ \cline{3-11} 
		\multicolumn{1}{|c|}{}                               &                                      & \multicolumn{1}{c|}{\textbf{Original}} & \multicolumn{1}{c|}{\textbf{3D Mask}} & \textbf{Protected} & \multicolumn{1}{c|}{\textbf{Original}} & \multicolumn{1}{c|}{\textbf{3D Mask}} & \textbf{Protected} & \multicolumn{1}{c|}{\textbf{Original}} & \multicolumn{1}{c|}{\textbf{3D Mask}} & \textbf{Protected} \\ \hline
		\multicolumn{1}{|c|}{\raisebox{1.5\height}{\begin{tabular}[c]{@{}c@{}}Bradley\\ Cooper\end{tabular}}}  &   \multicolumn{1}{c|}{\includegraphics[width=44pt]{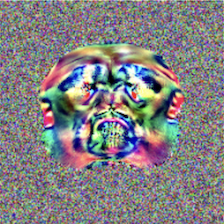}}                       & \multicolumn{1}{c|}{\includegraphics[width=44pt]{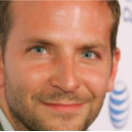}}                 &     \includegraphics[width=44pt]{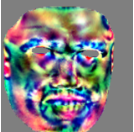}              & \multicolumn{1}{c|}{\includegraphics[width=44pt]{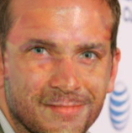}}                       & \multicolumn{1}{c|}{\includegraphics[width=44pt]{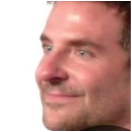}}                 &    \includegraphics[width=44pt]{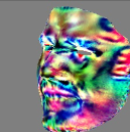}                & \multicolumn{1}{c|}{\includegraphics[width=44pt]{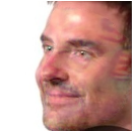}}                       & \multicolumn{1}{c|}{\includegraphics[width=44pt]{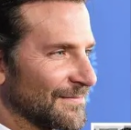}}                 &   \includegraphics[width=44pt]{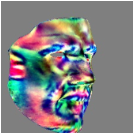}    & \includegraphics[width=44pt]{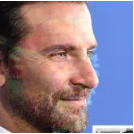}                                              \\ \hline
		\multicolumn{1}{|c|}{\raisebox{1.5\height}{\begin{tabular}[c]{@{}c@{}}Hugh\\ Grant\end{tabular}}}  &   \multicolumn{1}{c|}{\includegraphics[width=44pt]{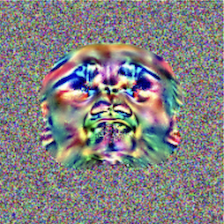}}                       & \multicolumn{1}{c|}{\includegraphics[width=44pt]{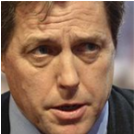}}                 &     \includegraphics[width=44pt]{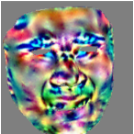}              & \multicolumn{1}{c|}{\includegraphics[width=44pt]{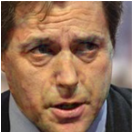}}                       & \multicolumn{1}{c|}{\includegraphics[width=44pt]{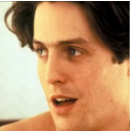}}                 &    \includegraphics[width=44pt]{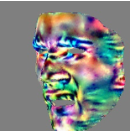}                & \multicolumn{1}{c|}{\includegraphics[width=44pt]{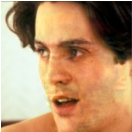}}                       & \multicolumn{1}{c|}{\includegraphics[width=44pt]{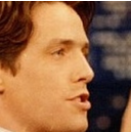}}                 &   \includegraphics[width=44pt]{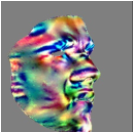}    & \includegraphics[width=44pt]{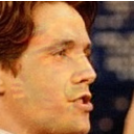}                                              \\ \hline
		\multicolumn{1}{|c|}{\raisebox{1.5\height}{\begin{tabular}[c]{@{}c@{}}Debra\\ Messing\end{tabular}}}  &   \multicolumn{1}{c|}{\includegraphics[width=44pt]{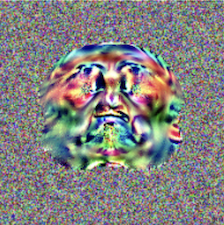}}                       & \multicolumn{1}{c|}{\includegraphics[width=44pt]{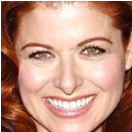}}                 &     \includegraphics[width=44pt]{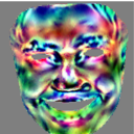}              & \multicolumn{1}{c|}{\includegraphics[width=44pt]{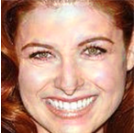}}                       & \multicolumn{1}{c|}{\includegraphics[width=44pt]{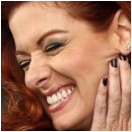}}                 &    \includegraphics[width=44pt]{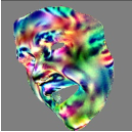}                & \multicolumn{1}{c|}{\includegraphics[width=44pt]{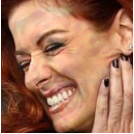}}                       & \multicolumn{1}{c|}{\includegraphics[width=44pt]{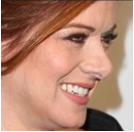}}                 &   \includegraphics[width=44pt]{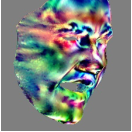}    & \includegraphics[width=44pt]{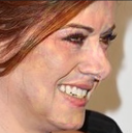}                                              \\ \hline
		\multicolumn{1}{|c|}{\raisebox{1.5\height}{\begin{tabular}[c]{@{}c@{}}Felicity\\ Huffman\end{tabular}}}  &   \multicolumn{1}{c|}{\includegraphics[width=44pt]{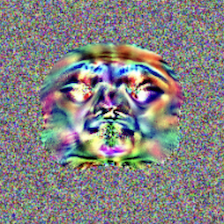}}                       & \multicolumn{1}{c|}{\includegraphics[width=44pt]{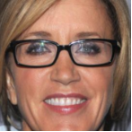}}                 &     \includegraphics[width=44pt]{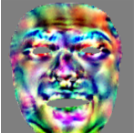}              & \multicolumn{1}{c|}{\includegraphics[width=44pt]{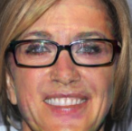}}                       & \multicolumn{1}{c|}{\includegraphics[width=44pt]{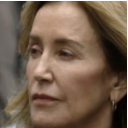}}                 &    \includegraphics[width=44pt]{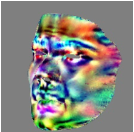}                & \multicolumn{1}{c|}{\includegraphics[width=44pt]{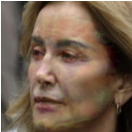}}                       & \multicolumn{1}{c|}{\includegraphics[width=44pt]{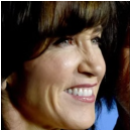}}                 &   \includegraphics[width=44pt]{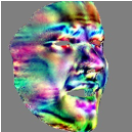}    & \includegraphics[width=44pt]{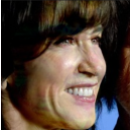}                                              \\ \hline
	\end{tabular}
	\caption{The same PPT (2nd column) can be used to protect any facial images of the same person. \scheme{} reads the original image (3rd, 6th, 9th columns), deforms the PPT into a 3D mask (4th, 7th, 10th columns), and generates the protected images (5th, 8th, 11th columns). The protected images look visually similar to their original counterparts.}\label{tab:static-photos}
\end{table*}
The privacy intruder’s FR-driven system is significantly less effective at retrieving relevant DB entries belonging to \scheme{} users. Figure~\ref{fig:psr-all} compares \scheme{} with two state-of-the-art methods, Chameleon~\cite{chow2024personalized} and OPOM~\cite{zhong2022opom}, across both datasets. To ensure a systematic evaluation, we assess their effectiveness under three scenarios spanning two levels of difficulty.

\noindent\textbf{Easy but Unrealistic Scenarios.}
\scheme{} can effectively disable retrieval when either all queries or all DB entries (but not both) are protected. It reduces the recall from $71.68\%$ to at most $1.05\%$ on FaceScrub, and from $70.09\%$ to at most $2.22\%$ on LFW. These settings are implicitly assumed in prior works, which explains why existing methods also perform well. We categorize them as easy scenarios because the protected image (either the query or DB entry) is explicitly optimized to differ from its unprotected counterpart. However, these settings are unrealistic for two key reasons. First, anti-FR solutions like \scheme{} are designed to protect facial images before they are uploaded online, so that the DB compiled through web scraping is likely to include at least some protected entries. Second, the privacy intruder might use protected images as queries, whether knowingly or not. A query image may be obtained from online sources and used to search for its digital footprint in the DB. In summary, although existing methods perform well under these conditions, such scenarios rarely occur in real-world settings.

\noindent\textbf{Hard but Realistic Scenarios.} \scheme{} remains effective even when both the queries and DB entries are protected, with the recall dropping from $71.68\%$ to $18.09\%$ on FaceScrub and from $70.09\%$ to $20.00\%$ on LFW. This is considered a hard scenario because protected images often form dense clusters (Figure~\ref{fig:prot-opt}), making it easier for a protected query to match protected DB entries. As a result, existing methods lead to only a minor degradation in retrieval performance. The results above assume that all DB entries are protected, which may not reflect practical situations. To examine this further, we analyze how recall changes as an increasing percentage of DB entries are protected in Figure~\ref{fig:leaked-portion} on FaceScrub. When the percentage is zero, the scenario reduces to the easy case discussed earlier. As more entries are protected, \scheme{} maintains a consistently low recall, indicating that the protected DB entries are unlikely to be retrieved. In contrast, existing methods exhibit a more significant increase in recall as protection coverage rises. Overall, \scheme{} achieves recall reductions that are $3.5\times$ and $2.7\times$ better than Chameleon and OPOM, respectively. Furthermore, \scheme{}’s effectiveness is consistent across different users. Figure~\ref{fig:user-psr} reports per-user recall, averaged over five random seeds, with standard deviations shown as error bars. \scheme{} consistently achieves the lowest recall with minimal variation, demonstrating both effectiveness and stability.

\noindent\textbf{Qualitative Studies.} Table~\ref{tab:matching-baselines} presents the top-5 retrieval results for the same query (1st column) under three protection methods (a–c) considering the hard case. Under \scheme{} (a), a protected query of B. Cooper incorrectly retrieves entries of Day-Lewise, J. Meyers, M. Lillard, and J. Cleese. In contrast, all retrieved entries under Chameleon (b) and OPOM (c) still belong to B. Cooper, as these methods generate protected images with high feature similarity.

Due to similar observations, we focus on FaceScrub and the hard case in the main paper.

\subsection{Visual Coherence in Photos and Videos}\label{sec:eval-natural}
\textbf{Static Photos.} \scheme{}'s 3D design preserves the visual quality of protected images well, as shown in Table~\ref{tab:static-photos}. For each user, we include their PPT (2nd column) and three example images to be protected (3rd, 6th, and 9th columns). The PPT is deformed according to each image to generate a 3D mask (4th, 7th, and 10th columns), which is then applied to the original image to produce the protected version (5th, 8th, and 11th columns). The protection of one image only takes 93 milliseconds on NVIDIA GeForce RTX 4090. These examples are chosen to reflect diverse head poses and facial expressions. The deformation aligns the protective pattern with the face well, maintaining visual consistency. 

\noindent\textbf{Videos.} Such a capability enables \scheme{} to deliver superior naturalness when protecting faces in videos. Demo videos are included on the GitHub repository, with three example frames shown in Table~\ref{tab:video-snapshots} (1st column), along with comparisons between \scheme{} (2nd and 3rd columns), Chameleon (4th column), and OPOM (5th column). \scheme{} achieves more natural-looking protection than Chameleon and OPOM, which expect the face to be front-facing, which rarely holds in practice, especially when protecting videos.

\begin{figure}[t]
\centering
\begin{subfigure}{0.326\linewidth}
	\centering
	\includegraphics[width=\linewidth]{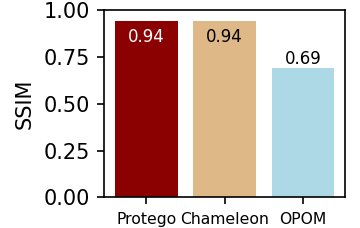}
	\caption{SSIM}\label{fig:visual-quality-ssim}
\end{subfigure}
\begin{subfigure}{0.326\linewidth}
	\centering
	\includegraphics[width=\linewidth]{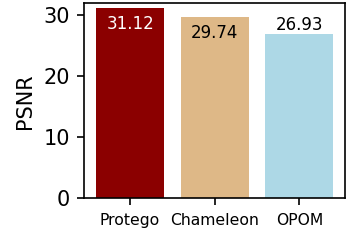}
	\caption{PSNR}\label{fig:visual-quality-psnr}
\end{subfigure}
\begin{subfigure}{0.326\linewidth}
\centering
\includegraphics[width=\linewidth]{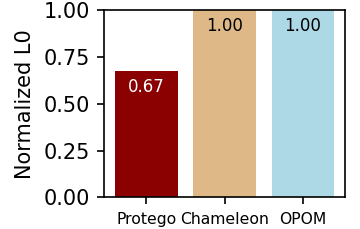}
\caption{Normalized L0}\label{fig:visual-quality-l0}
\end{subfigure}
\caption{\scheme{} maintains high visual quality in the protected images.}\label{fig:visual-quality}
\end{figure}

\noindent\textbf{Quantitative Studies.} \scheme{} outperforms existing methods in various image quality metrics, including SSIM, PSNR, and normalized L0 distance (Figure~\ref{fig:visual-quality}). \scheme{} achieves SSIM scores comparable to Chameleon and significantly outperforms OPOM. Its advantages are most notable in PSNR and normalized L0 norm, as its 3D design provides greater flexibility for preserving visual quality, and the protection is applied only to the facial region.

\begin{table}[t]\centering\small
	\setlength\tabcolsep{0.8pt}
	\textcolor{red}{\textbf{\textless Demo Videos Available on the GitHub repository\textgreater}}\\ \vspace{0.5em}
	\begin{tabular}{|c|cc|c|c|}
		\hline
		\textbf{\begin{tabular}[c]{@{}c@{}}Original\\ Video\end{tabular}} & \multicolumn{2}{c|}{\textbf{\begin{tabular}[c]{@{}c@{}}\scheme{}\end{tabular}}}  & \textbf{\begin{tabular}[c]{@{}c@{}}Chameleon\end{tabular}} & \textbf{\begin{tabular}[c]{@{}c@{}}OPOM\end{tabular}} \\ \hline
		 \includegraphics[width=46pt]{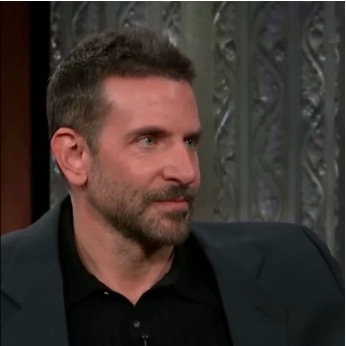}                                                               &     \includegraphics[width=46pt]{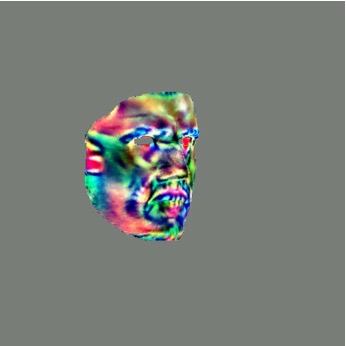}              &      \includegraphics[width=46pt]{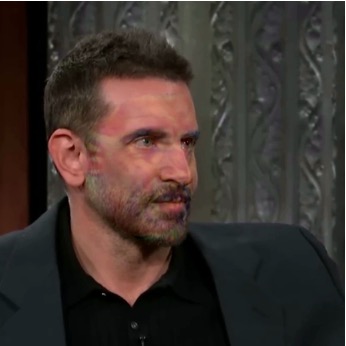}   &      \includegraphics[width=46pt]{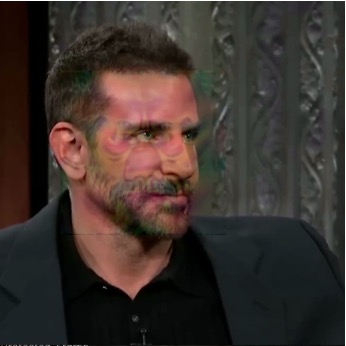}&      \includegraphics[width=46pt]{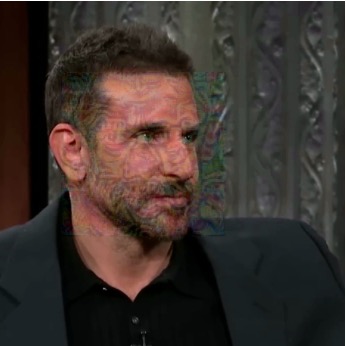}                                                            \\ \hline
		     \includegraphics[width=46pt]{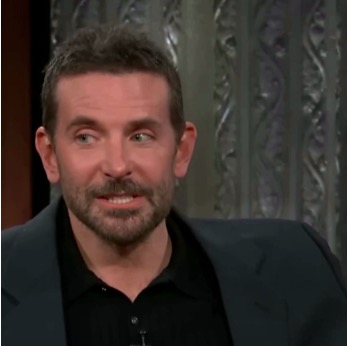}                                                               &     \includegraphics[width=46pt]{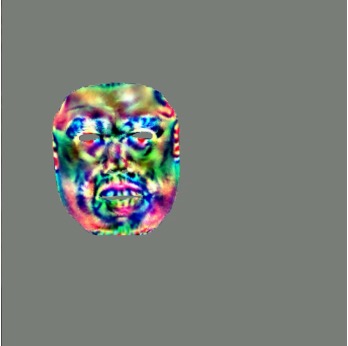}              &      \includegraphics[width=46pt]{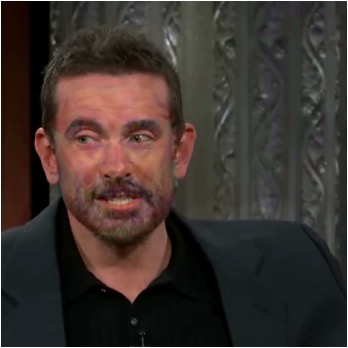}   &      \includegraphics[width=46pt]{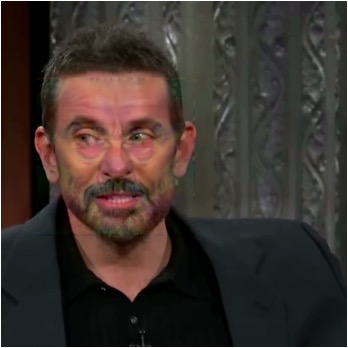} &      \includegraphics[width=46pt]{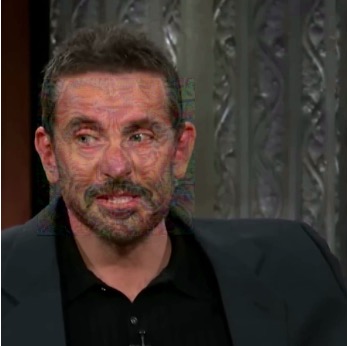}                                                             \\ \hline
		    \includegraphics[width=46pt]{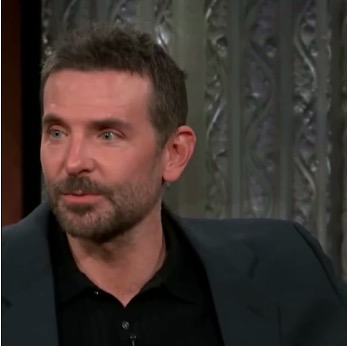}                                                               &     \includegraphics[width=46pt]{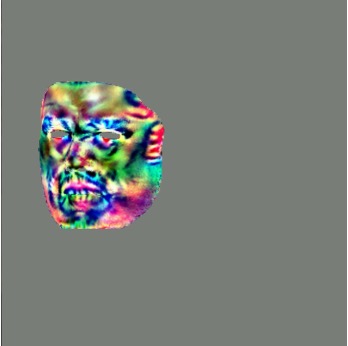}              &      \includegraphics[width=46pt]{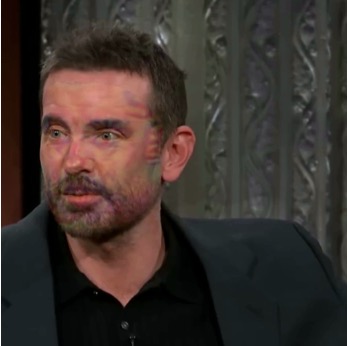}   &      \includegraphics[width=46pt]{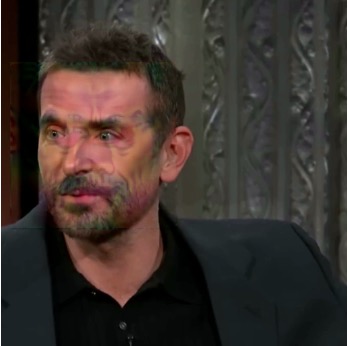}  &      \includegraphics[width=46pt]{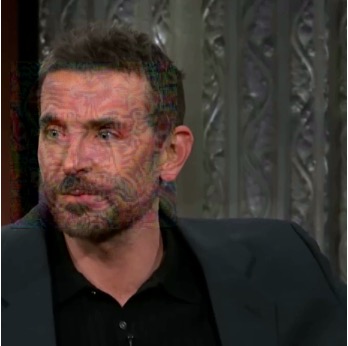}                                                              \\ \hline
	\end{tabular}
	\caption{Protego’s 3D design makes it well-suited for video protection, as the user’s face may appear differently across frames. It ensures consistency and naturalness. Please refer to the videos for easier comparison.}\label{tab:video-snapshots}
\end{table}

\subsection{Protect Against Different Unknown Models}\label{sec:eval-transfer}
\begin{figure}[t]
	\centering
	\includegraphics[width=0.95\linewidth]{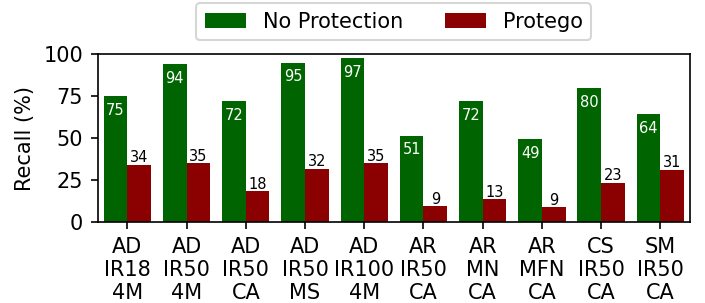}
	\caption{\scheme{} can protect against FR models not seen during the protection process.}\label{fig:transfer}
\end{figure}

\scheme{} can protect against FR models built using various methods (e.g., AdaFace~\cite{kim2022adaface}, ArcFace~\cite{deng2019arcface}), neural architectures (e.g., ResNet~\cite{he2016deep}, MobileNet~\cite{howard2017mobilenets}), and training datasets (e.g., CASIA-WebFace~\cite{yi2014learning}, MS1MV2~\cite{deng2019arcface}). We adopt a leave-one-out strategy to evaluate its effectiveness in Figure~\ref{fig:transfer} and observe a $2\times$ to $5\times$ reduction in recall under \scheme{}'s protection. Note that this represents the hard case where protected queries are used to retrieve protected DB entries.

A small team of diverse FR models can already suffice to generate a PPT with strong transferability. In particular, \scheme{} is compatible with focal diversity~\cite{wu2024hierarchical,chow2022boosting,chow2024diversity,chow2024personalized}, a lightweight method that quantifies the synergy of each combination of models from a given collection. We use focal diversity to find a 3-member team, train \scheme{} using the selected ensemble, and evaluate the protection against two FR models that are not in the collection (Table~\ref{tab:fr-collection}): MagFace~\cite{meng2021magface} and FaceNet~\cite{schroff2015facenet}. The team consists of AD-IR100-4M, AR-MN-CA, and SM-IR50-CA. Such a small team can already drive  \scheme{} to reduce the recall of MagFace from $98\%$ to $34\%$, and that of FaceNet from $82\%$ to $20\%$.

\begin{figure}[t]
\centering
	\includegraphics[width=0.88\linewidth]{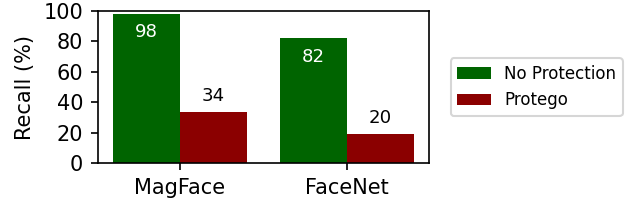}
	\caption{A small three-member team is sufficient to protect against unseen models like MagFace and FaceNet.}\label{fig:focal-diversity}
\end{figure}
\begin{figure}[t]
\centering
\includegraphics[width=\linewidth]{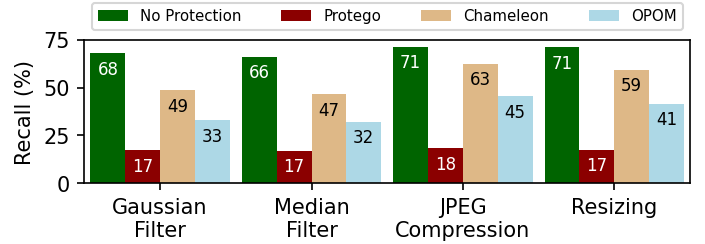}
\caption{Protego withstands various adaptive attacks launched by the privacy intruder to remove the protection perturbations.}\label{fig:adaptive-attacks}
\end{figure}

\subsection{Resilience Against Adaptive Privacy Intruder}
\scheme{} remains effective even when a privacy intruder attempts to remove potential protective perturbations from scraped images. Following \citet{chow2024personalized}, we apply Gaussian filtering, median filtering, JPEG compression, and resizing to protected faces. Figure~\ref{fig:adaptive-attacks} compares \scheme{} with Chameleon and OPOM. Aligned with the prior study, these operations may enhance the protection effect. \scheme{}'s recall remains largely unchanged. In contrast, while the other methods already struggle under the hard scenario, Chameleon's recall fluctuates between $49\%$ and $63\%$, and OPOM's between $33\%$ and $45\%$. We attribute \scheme{}'s resilience to the design of the PPT, which must be converted into a 3D mask via grid sampling, a process that inherently demands robustness to minor perturbations. While fine-tuning the FR model using protected facial images may weaken the protection, it requires manual labeling, which is impractical for large-scale search engines.

\subsection{Ablation Studies}
Each component contributes to the effectiveness of \scheme{}. The hypersensitivity loss plays the most critical role in reducing recall. Without this loss (i.e., without encouraging the protected images to exhibit diverse features), the recall jumps from $18.09\%$ to $78.52\%$. In fact, the privacy risk becomes even more severe than using no protection at all, which yields a recall of $71.68\%$. This counterintuitive result occurs because the protected images collapse into a dense cluster, making them easier to retrieve. Regarding the 3D protection design, we have previously shown its importance in producing natural-looking, high-quality protected images. In addition, it contributes to effectiveness, lowering the recall from $20.51\%$ to $18.09\%$, particularly by strengthening the protection on faces with extreme poses and expressions.

\section{Conclusions}
We have introduced \scheme{}, a user-centric, pose-invariant approach designed to minimize users' digital footprints. \scheme{} makes significant progress on two fronts. First, it prevents protected entries from being retrieved, even when privacy intruders use protected facial images as queries, which is a common and realistic threat scenario. Second, its pose-invariant design allows efficient adaptation to the pose and expression of any facial image, ensuring visually coherent protection that is especially crucial for video content. Overall, \scheme{} contributes to the fight against the misuse of FR for mass surveillance and unsolicited identity tracing.

\bibliography{aaai2026}

\end{document}